# Bayes and Naive-Bayes Classifier

Vijaykumar B (B091956), Vikramkumar (B092633), Trilochan (B092654)

*Computer Science & Engineering*

*Rajiv Gandhi University of Knowledge Technologies*
*Andhra Pradesh, India*

*Abstract* — Bayes and Naive-Bayes Classifier

## Introduction

The Bayesian Classification represents a supervised learning method as well as a statisticalmethod for classification. Assumes an underlying probabilistic model and it allows us to capture uncertainty about the model in a principled way by determining probabilities of the outcomes. It can solve diagnostic and predictive problems.

This Classification is named after Thomas Bayes ( 1702-1761), who proposed the Bayes Theorem. Bayesian classification provides practical learning algorithms and prior knowledge and observed data can be combined. Bayesian Classification provides a useful perspective for understanding and evaluating many learning algorithms. It calculates explicit probabilities for hypothesis and it is robust to noise in input data. In statistical classification the Bayes classifier minimises the probability of misclassification. That was a visual intuition for a simple case of the Bayes classifier, also called:

- Idiot Bayes
- Naïve Bayes
- Simple Bayes

We are about to see some of the mathematical formalisms, and more examples, but keep in mind the basic idea.Find out the probability of the previously unseen instance belonging to each class, then simply pick the most probable class.The structure of a Bayes classifier is determined by the conditional densities $p(x|\omega i)$ as well as by the prior probabilities.

A Bayes classifier is easily and naturally represented in this way. For the general case with risks, we can let $g_i(x) = -R(\alpha_i|x)$, since the maximum discriminant function will then correspond to the minimum conditional risk. For the minimum error rate case, we can simplify things further by taking $g_i(x) = P(\omega_i|x)$, so that the maximum discriminant function corresponds to the maximum posterior probability.

## Why Bayes Classification ?

**Probabilistic learning:** Calculate explicit probabilities for hypothesis, among the most practical approaches to certain types of learning problems.
**Incremental:** Each training example can incrementally increase/decrease the probability that a hypothesis is correct. Prior knowledge can be combined with observed data.
**Probabilistic prediction:** Predict multiple hypotheses, weighted by their probabilities.
**Standard:** Even when Bayesian methods are computationally intractable, they can provide a standard of optimal decision making against which other methods can be measured.

## Definition

The Bayesian Classification represents a supervised learning method as well as a statisticalmethod for classification. Assumes an underlying probabilistic model and it allows us to capture uncertainty about the model in a principled way by determining probabilities of the outcomes. It can solve diagnostic and predictive problems.A classifier is a rule that assigns to an observation a guess or estimate of what the unobserved label actually was. In theoretical terms, a classifier is a measurable function, with the interpretation that C classifies the point x to the class C(x). The probability of misclassification, or risk.

## Learning Classifiers based on Bayes Rule

Here we consider the relationship between supervised learning, or function approximation problems, and Bayesian reasoning. We begin by considering how todesign learning algorithms based on Bayes rule. Consider a supervised learning problem in which we wish to approximate an unknown target function f: X → Y , or equivalently P(Y|X). To begin, we will assume Y is a boolean valued random variable, and X is a vector containing n boolean attributes. In other words, $X = X_1, X_2 . . . , X_n$, where $X_i$ is the boolean random variable denoting the ith attribute of X.

Applying Bayes rule, we see that P(Y = yi |X) can be represented as

$$P(Y=y_i|X=x_k) = \frac{P(X=x_k|Y=y_i)P(Y=y_i)}{\sum_j P(X=x_k|Y=y_j)P(Y=y_j)}$$

where ym denotes the mth possible value for Y , xk denotes the kth possible vector value for X, and where the summation in the denominator is over all legal values of the random variable Y.

One way to learn P(Y |X) is to use the training data to estimate P(X|Y ) and P(Y ). We can then use these estimates, together with Bayes rule above, to determine P(Y |X = $x_k$ ) for any new instance $x_k$.

A note on Notation: We will consistently use upper case symbols (e.g., X) to refer to random variables, including both vector and non-vector variables. If X is a vector, then we use subscripts (e.g., $X_i$ to refer to each random variable, or feature, in X). We use lower case symbols to refer to values of random variables (e.g.,$X_i = x_{ij}$ may refer to random variable $X_i$ taking on its jth possible value). We will sometimes abbreviate by omitting variable names, for example abbreviating $P(X_i = x_{ij} |Y = y_k)$ to $P(x_{ij} |y_k)$. We will write E[X] to refer to the expected valueof X. We use superscripts to index training examples. We use δ(x) to denote an

"indicator" function whose value is 1 if its logical argument x is true, and whose value is 0 otherwise. We use the #D{x} operator to denote the number of elements in the set D that satisfy property x. We use a "hat" to indicate estimates; for example, $\hat{\Theta}$ indicates an estimated value of θ.

**Unbiased Learning of Bayes Classifiers is Impractical:**

If we are going to train a Bayes classifier by estimating P(X|Y) and P(Y), then it is reasonable to ask how much training data will be required to obtain reliable estimates of these distributions. Let us assume training examples are generated by drawing instances at random from an unknown underlying distribution P(X), then allowing a teacher to label this example with its Y value.

A hundred independently drawn training examples will usually suffice to obtain a maximum likelihood estimate of P(Y) that is within a few percent of its correct value when Y is a boolean variable. However, accurately estimating P(X|Y) typically requires many more examples. To see why, consider the number of parameters we must estimate when Y is boolean and X is a vector of n boolean attributes. In this case, we need to estimate a set of parameters

$$\theta_{ij} \equiv P(X = x_i | Y = y_j)$$

where the index i takes on $2^n$ possible values (one for each of the possible vector values of X), and j takes on 2 possible values. Therefore, we will need to estimate approximately $2^{n+1}$ parameters. To calculate the exact number of required parameters, note for any fixed j, the sum over i of θi j must be one. Therefore, for any particular value $y_j$, and the $2^n$ possible values of $x_i$, we need compute only $2^n - 1$ independent parameters. Given the two possible values for Y, we must estimate a total of $2(2^n - 1)$ such $\theta_{ij}$ parameters. Unfortunately, this corresponds to two distinct parameters for each of the distinct instances in the instance space for X. Worse yet, to obtain reliable estimates of each of these parameters, we will need to observe each of these distinct instances multiple times! This is clearly unrealistic in most practical learning domains. For example, if X is a vector containing 30 boolean features, then we will need to estimate more than 3 billion parameters.

# Naive Bayes Classifier

**Naive Bayes Classifier Introductory Overview:** The Naive Bayes Classifier technique is based on the so-called Bayesian theorem and is particularly suited when the Trees dimensionality of the inputs is high. Despite its simplicity, Naive Bayes can often outperform more sophisticated classification methods.

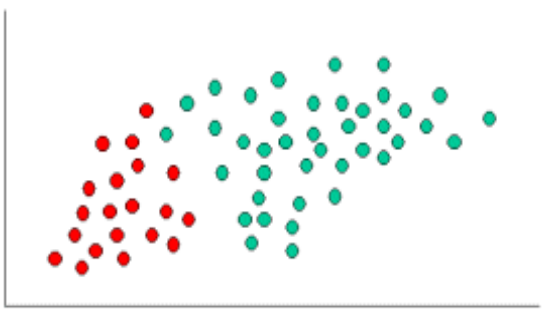

To demonstrate the concept of Naïve Bayes Classification, consider the example displayed in the illustration above. As indicated, the objects can be classified as either GREEN or RED. Our task is to classify new cases as they arrive, i.e., decide to which class label they belong, based on the currently exiting objects. Since there are twice as many GREEN objects as RED, it is reasonable to believe that a new case (which hasn't been observed yet) is twice as likely to have membership GREEN rather than RED. In the Bayesian analysis, this belief is known as the prior probability. Prior probabilities are based on previous experience, in this case the percentage of GREEN and RED objects, and often used to predict outcomes before they actually happen.

**Algorithm :**

Given the intractable sample complexity for learning Bayesian classifiers, we must look for ways to reduce this complexity. The Naive Bayes classifier does this by making a conditional independence assumption that dramatically reduces the number of parameters to be estimated when modeling P(X|Y), from our original $2(2^n - 1)$ to just 2n.

**Definition:** Given random variables X,Y and Z, we say X is conditionally independent of Y given Z, if and only if the probability distribution governing X is independent of the value of Y given Z; that is

$$(\forall i, j, k) P(X = x_i | Y = y_j, Z = z_k) = P(X = x_i | Z = z_k)$$

As an example, consider three boolean random variables to describe the current weather: Rain, Thunder and Lightning. We might reasonably assert that Thunder is independent of Rain given Lightning. Because we know Lightning causes Thunder, once we know whether or not there is Lightning, no additional information about Thunder is provided by the value of Rain. Of course there is a clear dependence of Thunder on Rain in general, but there is no conditional dependence once we know the value of Lightning.

**Derivation of Naive Bayes Algorithm :**

The Naive Bayes algorithm is a classification algorithm based on Bayes rule, that assumes the attributes $X_1 \ldots X_n$ are all conditionally independent of one another, given Y. The value of this assumption is that it dramatically simplifies the representation of P(X|Y), and the problem of estimating it from the training data. Consider, for example, the case where X = $X_1, X_2$. In this case

$$P(X|Y) = P(X_1, X_2 | Y)$$
$$= P(X_1 | X_2, Y) P(X2 | Y)$$
$$= P(X_1 | Y) P(X_2 | Y)$$

Where the second line follows from a general property of probabilities, and the third line follows directly from our above definition of conditional independence. More generally, when X contains n attributes which are conditionally independent of one another given Y, we have

$$P(X_1 \ldots X_n | Y) = \prod_{i=1}^{n} P(X_i | Y)$$

Notice that when Y and the Xi are boolean variables,

we need only 2n parameters to define $P(X_i = x_{ik} | Y = y_j)$ for the necessary i, j, k. This is a dramatic reduction compared to the $2(2^n - 1)$ parameters needed to characterize $P(X|Y)$ if we make no conditional independence assumption.

Let us now derive the Naive Bayes algorithm, assuming in general that Y is any discrete-valued variable, and the attributes $X_1 \ldots X_n$ are any discrete or real valued attributes. Our goal is to train a classifier that will output the probability distribution over possible values of Y, for each new instance X that we ask it to classify. The expression for the probability that Y will take on its kth possible value, according to Bayes rule, is

$$P(Y=y_i|x_1\ldots x_n) = \frac{P(Y=y_k)P(x_1\ldots x_n|Y=y_k)}{\sum_j P(Y=y_j)P(x_1\ldots x_n|Y=y_j)}$$

where the sum is taken over all possible values $y_j$ of Y. Now, assuming the $X_i$ are conditionally independent given Y, we can use equation (1) to rewrite this as

$$P(Y=y_i|X_1\ldots X_n) = \frac{P(Y=y_k)\prod_i P(X_i|Y=y_k)}{\sum_j P(Y=y_j)\prod_i P(X_i|Y=y_j)}$$

Equation (2) is the fundamental equation for the Naive Bayes classifier. Given a new instance $X^{new}=X_1\ldots X_n$, this equation shows how to calculate the probability that Y will take on any given value, given the observed attribute values of X new and given the distributions $P(Y)$ and $P(X_i|Y)$ estimated from the training data. If we are interested only in the most probable value of Y, then we have the Naive Bayes classification rule:

$$Y \leftarrow \arg\max_{y_k} \frac{P(Y=y_k)\prod_i P(X_i|Y=y_k)}{\sum_j P(Y=y_j)\prod_i P(X_i|Y=y_j)}$$

which simplifies to the following (because the denominator does not depend on $y_k$).

$$Y \leftarrow \arg\max_{y_k} P(Y=y_k)\prod_i P(X_i|Y=y_k)$$

## Summary

Despite the fact that the far reaching independence assumptions are often inaccurate, the naive Bayes classifier has several properties that make it surprisingly useful in practice. In particular, the decoupling of the class conditional feature distributions means that each distribution can be independently estimated as a one dimensional distribution. This helps alleviate problems stemming from the curse of dimensionality, such as the need for data sets that scale exponentially with the number of features. While naive Bayes often fails to produce a good estimate for the correct class probabilities, this may not be a requirement for many applications. For example, the naive Bayes classifier will make the correct MAP decision rule classification so long as the correct class is more probable than any other class. This is true regardless of whether the probability estimate is slightly, or even grossly inaccurate. In this manner, the overall classifier can be robust enough to ignore serious deficiencies in its underlying naive probability model. Other reasons for the observed success of the naive Bayes classifier.